\documentclass[10pt,twocolumn,letterpaper]{article}

\usepackage{cvpr}              

\usepackage{graphicx}
\usepackage{amsmath}
\usepackage{amssymb}
\usepackage{booktabs}
\usepackage{bm}

\usepackage{enumitem}  
\usepackage[10pt]{moresize}
\usepackage[section]{placeins} 

\usepackage[pagebackref,breaklinks,colorlinks]{hyperref}

\usepackage[capitalize]{cleveref}
\crefname{section}{Sec.}{Secs.}
\Crefname{section}{Section}{Sections}
\Crefname{table}{Table}{Tables}
\crefname{table}{Tab.}{Tabs.}

\def\cvprPaperID{4954} 
\def\confName{CVPR}
\def\confYear{2023}

\newcommand{\T}{\mathrm{T}}
\newcommand{\R}{\mathbb{R}}
\newcommand{\matr}[1]{\mathbf{#1}}
\renewcommand{\vec}[1]{\mathbf{#1}}
\renewcommand{\tilde}[1]{\ensuremath{\widetilde{#1}}}
\renewcommand{\hat}[1]{\ensuremath{\widehat{#1}}}

\DeclareMathOperator*{\argmin}{arg\,min}

\newcommand{\norm}[1]{\left\lVert#1\right\rVert}

\begin{document}
\title{
Controllable GAN Synthesis Using Non-Rigid Structure-from-Motion
}


\author{René Haas \qquad Stella Graßhof  \qquad Sami S. Brandt \\
IT University of Copenhagen, Denmark\\
{\tt\small \{renha,stgr,sambr\}@itu.dk}
}


\newpage

\maketitle
\begin{abstract}
In this paper, we present an approach for combining non-rigid structure-from-motion (NRSfM) with deep generative models,
and propose an efficient framework for discovering trajectories in the latent space of 2D GANs corresponding to changes in 3D geometry.
Our approach uses recent advances in NRSfM and enables editing of the camera and non-rigid shape information associated with the latent codes without needing to retrain the generator.       
This formulation provides an implicit dense 3D reconstruction as it enables the image synthesis of novel shapes from arbitrary view angles and non-rigid structure. 
The method is built upon a sparse backbone, where a neural regressor is first trained to regress parameters describing the cameras and sparse non-rigid structure directly from the latent codes.  
The latent trajectories associated with changes in the camera and structure parameters are then identified by estimating the local inverse of the regressor in the neighborhood of a given latent code. 
The experiments show that our approach provides a versatile, systematic way to model, analyze, and edit the geometry and non-rigid structures of faces. 
\end{abstract}


\section{Introduction}
\begin{figure}
\centering
\begin{flushleft}
\small 
\hspace{13pt} Original 
\hspace{17pt}  
-----------------------
Edits
-----------------------
\end{flushleft}
\vspace{-0.3cm}
\includegraphics[trim={0 0 35cm 0},clip,width=\linewidth]{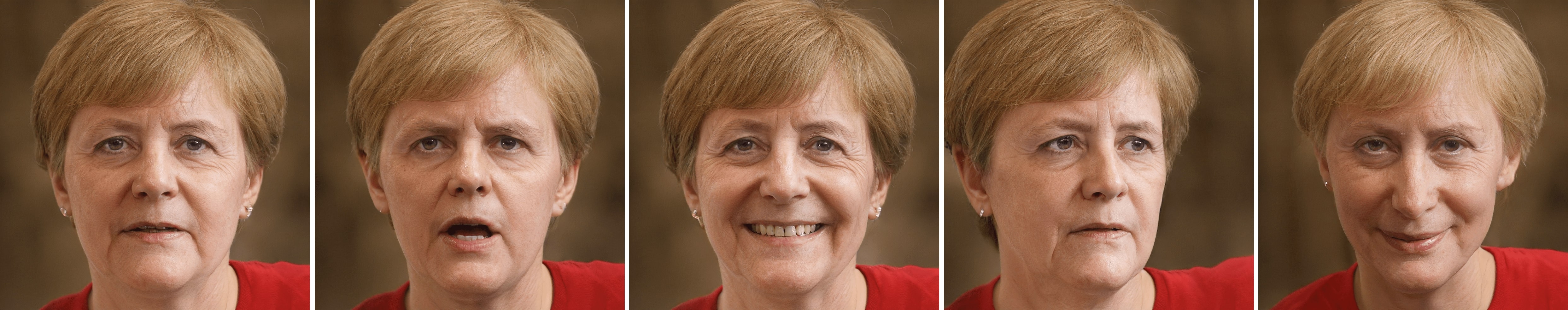}
\includegraphics[trim={0 0 35cm 0},clip,width=\linewidth]{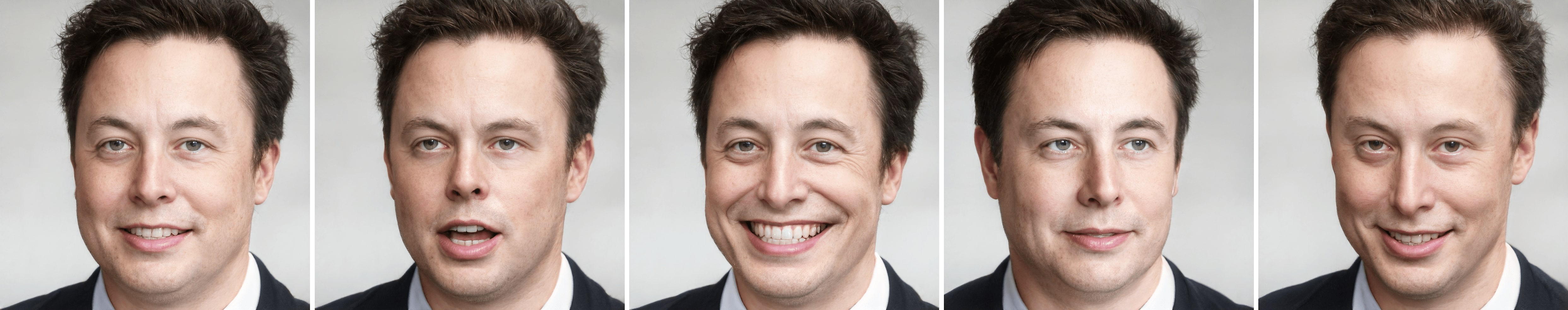}
\includegraphics[trim={0 0 35cm 0},clip,width=\linewidth]{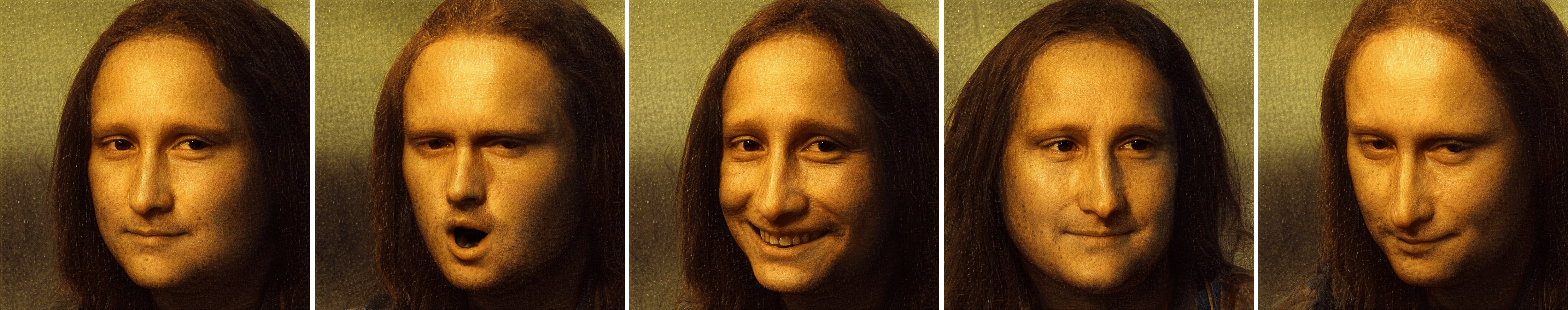}
\caption{ 
Our method parameterizes the latent space of an unconditional generator in terms of camera, orientation, and shape parameters. This allows for editing of rotation, translation, and non-rigid shape deformation of images created with StyleGAN. If coupled with a strong latent encoder, like e4e \cite{Tov2021e4e} or HyperStyle \cite{alaluf2021hyperstyle}, our method allows for semantic editing of real images.}
\label{fig:intro}
\vspace{0.3cm}
\end{figure} 
In recent years, Generative Adversarial Networks (GANs) \cite{Goodfellow2014GAN} have seen rapid improvements in image quality as well as training stability. 
GANs have achieved remarkable results in tasks such as image synthesis \cite{karras2018pggan,Karras2019StyleGAN, Karras2020StyleGAN2, Karras2020StyleGANada,Karras2021StyleGAN3}, 
image-to-image translation 
\cite{choi2020starganv2,richardson2021encoding, choi2018stargan}, semantic editing 
\cite{abdal2020img2sg++, Shen2020InterfaceganTPAMI, Harkonen2020GANSpace,Wu2020StyleSpace, patashnik2021styleclip, Tewari2020StyleRig, Abdal2020StyleFlow} as well as regression tasks \cite{nitzan2021large}.
Especially the StyleGAN \cite{Karras2019StyleGAN,Karras2020StyleGAN2,Karras2020StyleGANada,Karras2021StyleGAN3} family of models show state-of-the-art results in unconditional synthesis human faces images.
However, the standard StyleGAN architecture provides no way to directly control semantics like the pose and expression of the generated images. This has led to a large interest in finding semantic directions in the latent space of StyleGAN which controls specific semantic attributes such as pose, expression, hairstyle, illumination, etc.

\begin{figure}
\centering
\includegraphics[width = \linewidth]{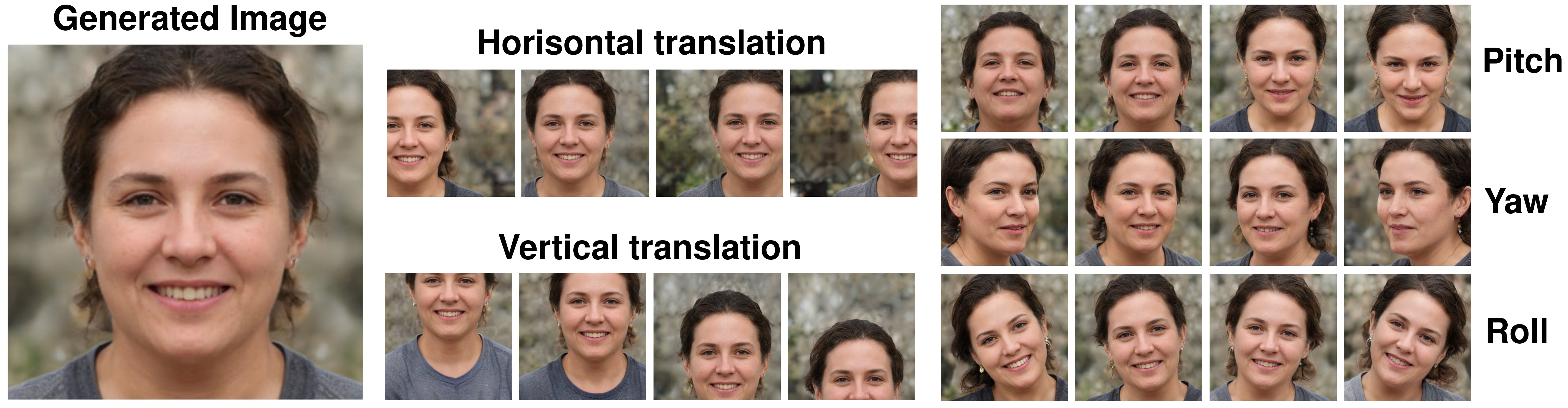}
\caption{Our method is able to discover trajectories in latent space corresponding to arbitrary rotations and translation. Note that translations and roll rotations are only possible when the generator has been trained on an unaligned data set, like the recent StyleGAN3 trained on FFHQU \cite{Karras2021StyleGAN3}.}
\label{fig:RotationTranslation}
\end{figure}

The non-rigid structure-from-motion (NRSfM) problem is a difficult, under-constrained problem with a long history in computer vision. NRSfM aims at obtaining the three-dimensional reconstruction of a scene with dynamical deformable structures from a sequence of 2D correspondences. Given a set of 2D correspondences, the standard assumption is that the deformable 3D shape is a linear combination of basis shapes; the camera information, describing how the 3D structure is projected onto the image plane, also needs to be recovered.
In this work, we incorporate a sparse 3D model based on NRSfM into a generative model like StyleGAN.
This is interesting for two reasons: first, this allows us to find trajectories in the latent space corresponding to well-defined semantic attributes corresponding to the camera geometry and non-rigid structure.
Second, using a generative model in conjunction with NRSfM provides a way to obtain an \emph{implicit} dense 3D reconstruction by using only the sparse 2D inputs.  
In other words, our approach enables dense image synthesis of novel shapes from arbitrary view angles and non-rigid deformation without the need for an explicit dense 3D reconstruction. An example of latent trajectories corresponding to changes in the camera parameters can be seen in Figure~\ref{fig:RotationTranslation}. 
Our method utilizes a sparse backbone that is a 3D model based on the approach for NRSfM given in  \cite{Brandt_Ackermann_2019,Grasshof_Brandt_2022}. The 3D model is constructed using solely 2D landmarks extracted from synthetic face images generated by StyleGAN, thus our approach required no 3D supervision. 

In this approach, we first factorize the measurement matrix, consisting of corresponding 2D landmark points, into a rigid and non-rigid part. Any arbitrary 3D shape can then be represented as the sum of a rigid basis shape and a linear combination of rank-one non-rigid basis shapes.
Our approach provides a way to recover a set of expansion coefficients that contains all the information about the 3D reconstruction of the extracted 2D face landmarks. 
Additionally, for each set of 2D landmarks, we recover a projection matrix, describing the camera information for projecting the 3D shapes onto the image plane as well as information about the orientation of the recovered 3D structure. 

We then proceed to connect the information recovered from the sparse 2D landmarks to the latent space of StyleGAN by training a regressor in the form of a multilayer perceptron (MLP) network to regress the shape and camera information directly from the latent codes. By estimating the local inverse of the regressor at a given latent code, we can identify trajectories in latent space corresponding to changes in camera or non-rigid geometry, while preserving other attributes of the generated image, like identity, texture, and illumination.
We show that the regressor network can be used for semantic manipulation of latent codes, either by using the first-order Taylor expansion of the trained network to define linear directions in latent space or by using the prediction of the network as a loss term for a gradient-based optimization algorithm.

As noted in \cite{wang2021hijackgan}, performing semantic editing in StyleGAN using only 2D landmarks is a very challenging problem since the 2D coordinates are extremely localized compared to more global attributes like age or gender.
In this work, we propose an editing framework that relies solely on 2D sparse landmarks. From the 2D landmarks, we use NRSfM to extract camera and shape parameters describing the 3D geometry. Our regressor then predicts these parameters images directly from the latent codes and we show how such a regressor naturally enables editing of the camera and non-rigid geometry of the synthesized images.

The main contributions of this paper are summarized as follows. 
\begin{itemize}
\itemsep0em 
\item We propose a framework that incorporates the NRSfM problem into the latent spaces of generative models.
\item Based on NRSfM we suggest a framework to get artistic control over images synthesized by StyleGAN. 
\item We show how our approach can model the camera, pose, and dense structure of the synthesized images, without an explicit dense 3D reconstruction. 
\item We propose a general method for enabling 3D awareness in 2D GANs without requiring any retraining or changes to the generator architecture. 
\item We propose a regularization technique that preserves the identity of the synthesized faces during the edits. 
\end{itemize}

\section{Related Work}
 
\vspace{0.25cm} \noindent
\textbf{StyleGAN.}
The StyleGAN \cite{Karras2019StyleGAN,Karras2020StyleGAN2,Karras2020StyleGANada,Karras2021StyleGAN3} generator is inspired by the style transfer literature \cite{Gatys2015StyleTransfer, Huang2017StyleTransfer} and consists of a  \emph{mapping network} $f$ which maps a latent vector $\mathbf{z}\in \mathcal{Z}$, sampled from the standard Gaussian $\mathcal{N}(\mathbf{0},\mathbf{I})$ in order to obtain an intermediate representation $\mathbf{w}\in \mathcal{W}$. 
The latent space $\mathcal{W}$ is more disentangled than $\mathcal{Z}$ \cite{Karras2019StyleGAN}. 
To synthesize an image, the latent code $\mathbf{w}$ is copied and fed to each synthesis block of the \emph{synthesis network} $G$ which produces the final image.
Instead of feeding the same vector to each of the synthesis blocks, if the vectors are allowed to differ, the resulting space is typically denoted as $\mathcal{W}+$. It has been shown that using $\mathcal{W}+$ space can lead to lower reconstruction loss when performing GAN inversion \cite{Abdal2019Image2StyleGAN, Zhu2020InDomain}, however at the cost of lower editability \cite{Tov2021e4e} of the resultant latent codes.

\vspace{0.25cm} \noindent
\textbf{Semantic Editing.}
Several methods have been proposed to enable semantic edits of the images produced by StyleGAN. 
InterFaceGAN \cite{Shen2020Interfacegan,Shen2020InterfaceganTPAMI} enables editing of binary semantic attributes like left/right pose, gender, presence or absence of smile, etc.
Here, a set of latent codes are first sampled and the images are annotated using pre-trained binary classifiers. Following the annotation step, a support vector machine was fitted on the labeled data for each binary semantic attribute. The normal vector for the supporting hyperplane then defines the semantic direction in latent space. 
Another approach for semantic editing is GANSpace \cite{Harkonen2020GANSpace} which proposes to use PCA on sampled latent codes to find semantic directions in an unsupervised fashion. Another related approach also factorizes the weights of the trained generator \cite{Shen2020SeFa} rather than the latent codes. Both methods then change the semantics of the generated images by perturbing latent codes in the direction of the found semantic directions.  
Additionally, \cite{Abdal2020StyleFlow} uses normalizing flows for attribute-conditioned semantic editing and explores both linear and non-linear trajectories in latent space. 
Related to our approach we also find StyleRIG \cite{Tewari2020StyleRig} which facilitates semantic editing in StyleGAN using 3D morphable models \cite{Blanz1999MorphableModel}. 
Recently it was proposed to regard the space of channel-wise style parameters after the learned affine transformation in each block in the StyleGAN synthesis network as a separate latent space, complementing the previously mentioned $\mathcal{Z}$, $\mathcal{W}$ and $\mathcal{W}+$ spaces. 
This latent space was named StyleSpace and denoted as $\mathcal{S}$ \cite{Wu2020StyleSpace}. It has been shown that $\mathcal{S}$ space has superior disentanglement properties, especially in StyleGAN3 \cite{Karras2021StyleGAN3,Alaluf2022SGThirdTime}, compared to $\mathcal{W}$ space thus enabling fine-grained and highly localized edits, like the closing of the eyes or changes to hair color \cite{Wu2020StyleSpace}. 

\vspace{0.25cm} \noindent
\textbf{Inversion.}
For purposes involving the editing of real images, it is necessary to find a good latent representation. That is, we need to find a latent code that, when passed to the generator, faithfully reconstructs the target image. This problem is known as GAN inversion.
Techniques for GAN inversion have typically either used an optimization-based approach, where the latent code is directly optimized in order to minimize the error (typically MSE or LPIPS \cite{Zhang2018LPIPS}) between the generated and target image \cite{Abdal2019Image2StyleGAN,abdal2020img2sg++,Karras2020StyleGAN2} or an encoder based approach, where a given target image is directly mapped into the latent space \cite{richardson2021encoding, Perarnau2016Invertible,alaluf2021restyle} or a hybrid approach \cite{Zhu2020InDomain,pbaylies}. 

Recent work \cite{Tov2021e4e} suggests that there is a trade-off between distortion and editability when selecting which latent space to project a given target image into. Projecting images into extended $\mathcal{W}+$ space typically leads to higher reconstruction quality \cite{Abdal2019Image2StyleGAN}, i.e., produces an image closer to the target images. However, latent codes in $\mathcal{W}+$ are generally less editable than latent codes in native $\mathcal{W}$ space. 

The e4e encoder proposed in \cite{Tov2021e4e} seeks to find a good trade-off between reconstruction and editability by projecting images into $\mathcal{W}+$ but constraining the latent codes to be close to $\mathcal{W}$. 
Recently \cite{Roich2021pivotal} shows that real images can be embedded into $\mathcal{W}$ space by fine-tuning the trained generator around the target image, thus circumventing the need for projecting into $\mathcal{W}+$ space.
In \cite{alaluf2021restyle}, a combination of the iterative and encoder-based methods is proposed. Here the encoder predicts the residual with respect to the current estimate of the latent code and thus is able to refine the latent code using only a few forward passes of the encoder in a process referred to as iterative refinement. 
Recently, \cite{alaluf2021hyperstyle} proposed to unite the ideas of fine-tuning the generator from \cite{Roich2021pivotal} with the iterative refinement from \cite{alaluf2021restyle} by introducing a hypernetwork which predicts how the parameters of the generator should be changed in order to faithfully embed a given real image into the native, and more editable, $\mathcal{W}$ space. 

\vspace{0.25cm} \noindent
\textbf{Explicitly 3D aware GANs.}
Several works have investigated incorporating explicit 3D understanding into GANs \cite{gu2021stylenerf,Niemeyer2021giraffe,ylmaz2022liftedgan}.
Compared to these, our approach can be used to control the 3D structure in existing 2D GANs without the need for adaptation of the generator architecture nor does our approach require any retraining.

\vspace{0.25cm} \noindent
\textbf{NRSfM.} 
Structure-from-motion (SfM) deals with the problem of inferring the scene geometry and camera information from image sequences. 
In \cite{Tomasi_Kanade_1992} an orthographic camera model was assumed to infer rigid shape and motion by a factorization of the measurement matrix. 
In \cite{Bregler_Hertzmann_Biermann_2000}, this problem was formulated to include non-rigid deformations by assuming that a shape is a linear combination of 3D basis shapes, hence proposing an approach for non-rigid structure-from-motion (NRSfM). 
Various works have followed up on this approach over the years this is still an area of active research \cite{bue2021nrsfmbenchmark}. 

Recently, there have been attempts to solve the NRSfM problem by employing neural networks. 
However, most require a large training data set \cite{KongLucey2019DeepNRSFM}, 3D supervision or an assumption of an orthographic camera model \cite{KongLucey2019DeepNRSFM,sidhu_neural_2020}. 
Specifically, \cite{KongLucey2019DeepNRSFM} formulates the NRSfM problem as a multi-layer block sparse dictionary learning problem converted into a deep neural network. 
%
In neural NRSfM \cite{sidhu_neural_2020}, the authors rely on dense 2D point tracks to recover dense 3D representations, and train an auto-decoder-based model with subspace constraints in the Fourier domain.
Our method differs from these works in several aspects, because (1) it relies only on sparse 2D points, (2) it does not rely on a block structure, and (3) it assumes an affine camera model. This makes our approach direct, lightweight, fast and efficient.

%



\section{Method}\label{sec:method}
\begin{figure*}[tb]
\centering
\includegraphics[width=\textwidth]{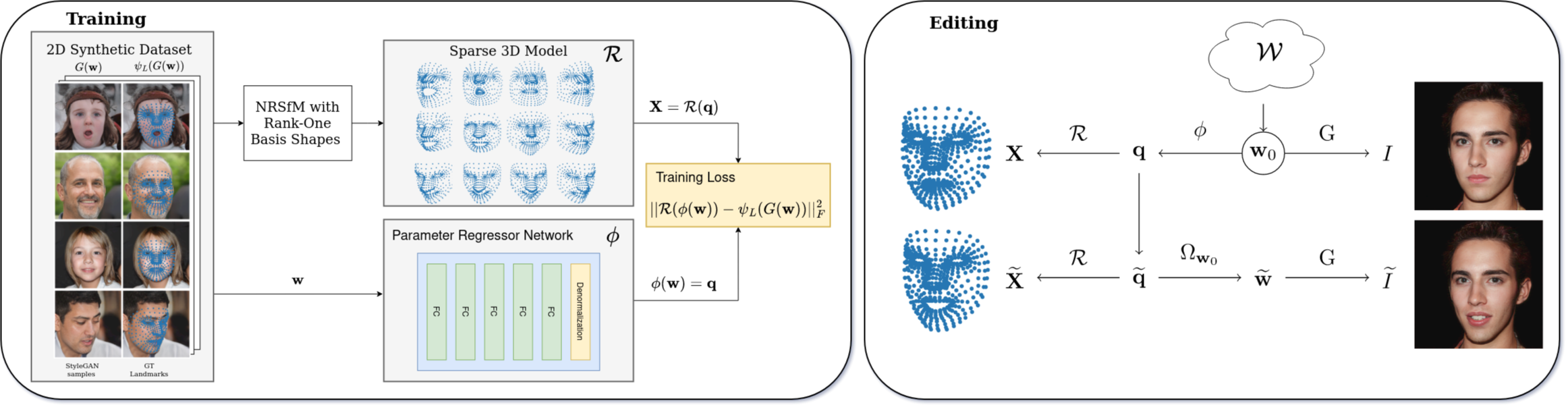}
\caption{ 
We first create a sparse 3D model $\mathcal{R}$ of facial landmarks from a data set of 2D landmarks $\mathbf{X}$ using NRSfM. The 3D model is parameterized by an attribute vector $\mathbf{q}$ which contains information about the camera, rotation, and non-rigid 3D structure.  
We then train a regressor $\phi$ to predict the parameters $\mathbf{q}$ directly from latent codes $\mathbf{w}$. Once the regressor is trained, it can be used for semantic editing. Given a latent code $\mathbf{w}_0$ with corresponding attribute vector $\mathbf{q}_0$ we can define a different, target attribute vector $\tilde{\mathbf{q}}$ and transfer it onto $\mathbf{w}_0$ using the transformation $\Omega_{\mathbf{w}_0}$ which depends on the regressor $\phi$. }
\label{fig:method}
\end{figure*}

Let $I=G(\mathbf{w})$ be an image generated by the StyleGAN generator by the latent code $\mathbf{w}$. Our goal is to locally parameterize the manifold of latent codes, in the neighborhood of a fixed latent code $\mathbf{w}_0$, by an \emph{attribute vector} $\mathbf{q}$ so that 
\begin{align}
\mathbf{w}=\Omega_{\mathbf{w}_0}(\mathbf{q}),
\end{align} 
where $\mathbf{q}$ describes the \emph{pose}, \emph{shape}, and \emph{camera} information of the generated image.
This formulation facilitates the transfer of the target attributes ${\mathbf{q}}$ onto the latent code $\mathbf{w}_0$ to obtain an edited code $\mathbf{w}$ where only the target attributes have changed in the image, while preserving all other attributes such as identity, textures, illumination, and hairstyle.

Our method is composed of three distinct elements.
(1) The \emph{sparse back-bone} relies on a pre-trained landmark extractor $\psi_L$, which extracts the 2D landmarks $\mathbf{X}=\psi_L(I)$, from a generated image $I$; and a closed-form parameterization 
for the 2D landmarks as $\mathbf{X} = \mathcal{R}(\mathbf{q})$, where $\mathcal{R}$ maps the 3D shape defined by the attribute vector $\mathbf{q}$ onto the image plane.
(2) The \emph{attribute regressor} $\phi$ predicts the attribute vector $\mathbf{q}=\phi(\mathbf{w})$ from the latent code $\mathbf{w}$, where the regressor is trained by minimizing by the square distance between the ground truth landmarks $\mathbf{X}=(\phi_L \circ G)(\mathbf{w})$ and predicted landmarks $\hat{\mathbf{X}}=(\mathcal{R} \circ \phi) (\mathbf{w})$, respectively. (3) The \emph{regression inversion} constructs the local inverse model of the regressor $\phi$ around the latent code $\mathbf{w}_0$, i.e., finds the local parameterization of the latent space so that $\mathbf{w}=\Omega_{\mathbf{w}_0}(\mathbf{q})$, where $\phi (\mathbf{w}_0)=\mathbf{q}_0$.


The remaining part of this section is organized as follows. In Section \ref{subsec:rankonemodel} we introduce the landmark parameterization $\mathcal{R}(\mathbf{q})$ and detail how the 3D basis shapes can be recovered from a data set of sparse 2D landmarks. The training of the attribute network $\phi$ is discussed in Section \ref{sec:rankoneregressor} and finally, in Section \ref{subsec:editing} we detail how the regressor $\phi$ is used to facilitate highly interpretable semantic editing.

\subsection{Rank-one model}\label{subsec:rankonemodel}
The rank-one approach for non-rigid structure from motion, proposed in \cite{Brandt_Ackermann_2019,Brandt_Ackermann_Grasshof,Grasshof_Brandt_2022}, is an affine camera model for non-rigid structure from motions which is able to recover 3D structure from sparse 2D correspondences using rank-one basis shapes.
In this paper, we frame the model as a parameterization of the space of possible 2D shapes in terms of camera, rotation, translation, and shape parameters. 
We propose to write the model of \cite{Brandt_Ackermann_2019,Brandt_Ackermann_Grasshof} in closed-form as
\begin{align}
\mathcal{R}(\mathbf{q})
&=\underbrace{\mathbf{K}[\mathbf{I}_2|\mathbf{0}]\mathbf{R}(\bm{\theta})}_{\matr{M}}
\left[\matr{B}_0 + \sum_{k = 1}^K \alpha_{k}\matr{B}_k\right]+\vec{t}\otimes \vec{1}_L^\T,
\label{eq:rankonemodel}
\end{align}
where $\mathbf{K}\in \mathbb{R}^{2\times 2}$ an upper triangular matrix, containing the camera parameters $\mathbf{k}=(k_{11},k_{12},k_{22})$.
The rotation matrix $\mathbf{R}\in\mathbb{R}^{3\times 3}$ is here parameterized in terms of the Euler angles $\bm{\theta}=(\theta_x,\theta_y,\theta_z)$.
The rigid basis shape $\mathbf{B}_0$ describes the average 3D reconstruction while the non-rigid basis shapes $\mathbf{B}_{k}$, for $k>0$ describe the non-rigid variation from the rigid basis shape.
The expansion coefficients $\bm{\alpha}$ determine the strength of the contribution of each of the non-rigid basis shapes $\mathbf{B}_{k}$. 
Finally, the translation vector $\mathbf{t}$ determines the offset from the origin. 
In \eqref{eq:rankonemodel}, $\otimes$ denotes the Kronecker product, $\vec{1}_L\in\R^L$ is a vector of ones, thus $\vec{t}\otimes \vec{1}_L^\T\in\R^{2\times L}$ yields a matrix where $\vec{t}\in\R^2$ is repeated $L$-times column-wise. 
To summarize, with \eqref{eq:rankonemodel} any 2D shape $\mathbf{X}$ can be parameterized in terms of an attribute vector $\mathbf{q}$ as $\mathbf{X} = \mathcal{R}(\mathbf{q})$ where the attribute vector contains the camera, rotation, shape, and translation parameters as $\mathbf{q} = (\mathbf{k},\bm{\theta},\bm{\alpha},\mathbf{t})$. 

In \eqref{eq:rankonemodel}, the only parameters which are not directly specified by the attribute vector are the rigid and non-rigid basis shapes $\mathbf{B}_0$ and $\matr{B}_k$, for $k>0$, respectively. 
In the next section, we will see how the rigid and non-rigid basis shapes can be recovered given a data set of corresponding 2D landmark points. 

\subsubsection{Non-rigid Factorization.}
Given a data set of $N$ 2D shapes $\matr{X}_n \in \R^{2\times L}$, 
we initially stack the 2D shapes $\matr{X}_n$, $n=1,2,\ldots,N$ into a measurement matrix $\mathcal{X} \in \R^{2N \times L}$. 
Our aim is to factorize $\mathcal{X}$ into a rigid $\matr{X}_0$ and non-rigid $\delta\matr{X}$ part such that 
\begin{align}
    \mathcal{X} = \mathcal{X}_0 + \delta\mathcal{X} = \matr{M}_0\matr{B}_0 + \delta \matr{M}\delta\matr{B}. 
\end{align}
To recover the rigid basis shape $\mathbf{B}_0$ from \eqref{eq:rankonemodel} we first calculate the singular value decomposition (SVD) of the measurement matrix as $\mathcal{X} = \mathbf{U}\bm{\Lambda}\mathbf{V}^\mathrm{T}$. The rigid part $\mathcal{X}_0$ is then constructed by selecting the three dominant singular vectors such that
\begin{align}
\mathcal{X}_0 
&= \mathbf{U}_0\mathbf{\Lambda}_0\mathbf{V}_0^\mathrm{T} 
=  \matr{M}_0 \matr{B}_0 \quad
\text{with} \\ 
\matr{M}_0 &= \mathbf{U}_0 \mathbf{\Lambda}_0 \in \mathbb{R}^{2N\times 3}, 
\quad 
\matr{B}_0 = \mathbf{V}_0^\mathrm{T} \in  \mathbb{R}^{3\times L}.
\end{align}
%
The matrix $\mathbf{M}_0$ contains the $N$ affine projection matrices $\matr{M}_n$, associated with each shape in the data set, which are stacked on top of each other in $\matr{M}_0$. 

To recover the non-rigid basis shapes $\matr{B}_k$, we subtract the rigid part from the measurement matrix, i.e., $\delta\mathcal{X} = \mathcal{X} - \mathcal{X}_0$, and calculate the SVD of the remaining part as 
\begin{align}
    \delta\mathcal{X} = \delta\mathbf{U}\delta\matr{\Lambda}\delta\mathbf{V}^T = \delta \matr{M} \delta \matr{B}. 
\end{align}
In the following, we use $\delta\matr{B} = \delta\mathbf{V}^\T \in \R^{L\times L}$ to construct the non-rigid basis shapes as $\matr{B}_k = \vec{d}_k \vec{b}_k^\T$, where $\vec{b}_k^\mathrm{T}$ is the $k$th row of $\delta\matr{B}$, and $\vec{d}_k$ is a $3 \times 1$ unit vector which will be determined by gradient based optimization. 
Now our goal is to recover $\matr{D} = [\vec{d}_1,\cdots,\vec{d}_K ]\in \mathbb{R}^{3\times K}$ which defines the non-rigid basis shapes. %
In \cite{Brandt_Ackermann_2019,Brandt_Ackermann_Grasshof,Grasshof_Brandt_2022}, $\matr{D}$ was recovered by an alternating least squares optimization scheme by exploiting the orthonormality of the non-rigid basis shapes. 
Here we use gradient-based optimization instead. For this purpose, it is convenient to  write the factorization of the measurement matrix $\mathcal{X}$ as 
\begin{align}
\mathcal{X}=\matr{M}_0\matr{B}_0 + \matr{M}^\alpha \matr{B},
\end{align}
with
\begin{align}
\begin{split}
\matr{M}^\alpha &=
(\bm{\alpha} \otimes \mathbf{1}_{2 \times 3}) \odot (\mathbf{1}_K \otimes \matr{M}_0) \\
&=
\begin{bmatrix}
\alpha_{11}\matr{M}_1 &\alpha_{12}\matr{M}_1 &\cdots & \alpha_{1K} \matr{M}_1  \\
\alpha_{21}\matr{M}_2&\alpha_{22}\matr{M}_2&\cdots& \alpha_{2K} \matr{M}_2  \\
\vdots & \vdots &\ddots & \vdots \\
\alpha_{N1}\matr{M}_N&\alpha_{N2}\matr{M}_N&\cdots& \alpha_{NK} \matr{M}_N
\end{bmatrix}        
\end{split},
\end{align}
where $\odot$ is the Hadamard product and 
\begin{align}
    \matr{B} = 
    \text{diag}(\text{vec}(\vec{D})) (\mathbf{I}_K \otimes \mathbf{1}_3) \delta\matr{B} = 
    \begin{bmatrix}
    \vec{d}_1\vec{b}_1^\T \\
    \vec{d}_2 \vec{b}_2^\T\\
    \vdots \\
    \vec{d}_K\vec{b}_K^\T
    \end{bmatrix} = 
    \begin{bmatrix}
        \matr{B}_1 \\
        \matr{B}_2 \\
        \vdots \\
        \matr{B}_K
        \end{bmatrix}.
        \label{eq:definition-basisshape}
\end{align}
Then we can jointly find $\matr{D}$ and $\bm{\alpha}$ by minimizing
\begin{align}
    \min_{\matr{D},\bm{\alpha}} ||\hat{\mathcal{X}}(\matr{D},\bm{\alpha}) - \mathcal{X} ||^2_F + \lambda \sum_{k = 1}^K (\vec{d}_k^\T \vec{d}_k - 1)^2, ~\lambda \in\R^+,
    \label{eq:minimization}
\end{align}
by gradient descent. Once we have found the $\matr{D}$ and $\bm{\alpha}$ which minimizes \eqref{eq:minimization}, the non-rigid basis shapes can be constructed using \eqref{eq:definition-basisshape}.
The found basis shapes $\mathbf{B}_i$ completely specify the parameterization in \eqref{eq:rankonemodel}.  

The parameterization of a new unseen set of landmarks $\mathbf{X}_\text{new}$ can be obtained as
\begin{align}
    \mathbf{q}^* = \argmin_{\mathbf{q}}  ||\mathcal{R}(\mathbf{q}) - \mathbf{X}_\text{new} ||^2_F.
    \label{eq:qrecover}
\end{align}


\subsection{Connection to the latent space} \label{sec:rankoneregressor} 
Having found the parameterization $\mathcal{R}$ in \eqref{eq:rankonemodel}, we train a MLP network $\phi$ to regress the parameters $\mathbf{q}$ directly from the latent codes $\mathbf{w}$ such that $\phi (\mathbf{w}) = \hat{\mathbf{q}}$. Predicting $\mathbf{q}$ is equivalent to predicting the landmarks of the generated images as $\mathcal{R}( \phi (\mathbf{w})) = \hat{\matr{X}}$. 
We train the network $\phi$ to minimize the objective function
\begin{align}
\mathcal{L}\left(\mathbf{w}\right) 
= \norm {\mathcal{R}( \phi (\mathbf{w}))  - \psi_L(G(\mathbf{w})) }^2_F,
\end{align} 
where $\psi_L$ is some pre-trained landmark extractor. 

\subsection{Semantic Editing} \label{subsec:editing}
In the following, we provide an analytic as well as a gradient-based approach for locally inverting the trained network $\phi$, to control the pose and non-rigid shape of images generated by StyleGAN.
%
For the \emph{analytic approach}, the first order Taylor expansion of $\phi$ around $\mathbf{w}_0$ yields 
\begin{align}
    \phi (\mathbf{w}) = \phi (\mathbf{w}_0) +  \mathbf{J}|_{\mathbf{w}= \mathbf{w}_0} (\mathbf{w}- \mathbf{w}_0),
\end{align}
where $\mathbf{J}|_{\mathbf{w}= \mathbf{w}_0}$ is the Jacobian of $\phi$ evaluated at $\mathbf{w}_0$. Now since $\phi (\mathbf{w}_0) = \mathbf{q}_0$ we can rewrite this as 
\begin{align}
    \mathbf{w} = \mathbf{w}_0 + \mathbf{J}^\dagger(\vec{q}-\vec{q}_0),
    \label{eq:linedit}
\end{align}
where $\mathbf{J}^\dagger$ is the Moore-Penrose pseudo-inverse of $\mathbf{J}|_{\mathbf{w}= \mathbf{w}_0}$. 
Thus we have a way of editing a latent code $\mathbf{w}_0$ which has associated 2D landmarks $\matr{X}_0$ parameterized by $\vec{q}_0$ as $ \matr{X}_0 = \mathcal{R}(\vec{q_0})$ in such a way as to obtain a new latent code $\vec{w}$ with the corresponding landmark parameterized by $\vec{q}$.

The analytic method described in \eqref{eq:linedit} requires evaluating $\mathbf{J}$ at $\mathbf{w}_0$ and defines a linear path in latent space. As an alternative to the analytic approach in \eqref{eq:linedit} we propose a \emph{gradient-based approach} where we directly minimize the difference between a target $\mathbf{q}_\text{target}$ and the network prediction $\phi (\mathbf{w})$ as
    \begin{align}
       \min_{\mathbf{w}} \norm{\phi(\mathbf{w}) - \mathbf{q}_\text{target}}^2 + \lambda \mathcal{D}(G(\mathbf{w}),G(\mathbf{w}_0)).
      \label{eq:gradientedit}
    \end{align}
Where $\mathcal{D}(\cdot , \cdot)$ is some image similarity metric such as LPIPS \cite{Zhang2018LPIPS} or Arcface \cite{Deng2019ArcFace}, which we employ for regularization purposes.
The gradient-based editing is analogous to what is proposed in \cite{wang2021hijackgan}. However, here we allow for the passing of gradients through the generator $G$ in order to calculate the identity loss in \eqref{eq:gradientedit}.

\subsection{Manifold Considerations}
Recent work has explored the manifolds of GANs in the context of Riemannian geometry \cite{Shao2018RiemannDGM,wang2021GANGeometry}.
Given a distance metric $\mathcal{D}(\cdot, \cdot)$, 
on the image space, we can define the distance $d$ between the latent codes $\mathbf{w}_1,\mathbf{w}_2$ as the distance between the images they generate in terms of the image metric $\mathcal{D}$, i.e. $d(\mathbf{w}_1,\mathbf{w}_2) = \mathcal{D}(G(\mathbf{w}_1),G(\mathbf{w}_2))$.
The Hessian $\mathbf{H}$ of the squared distance function can be viewed as the metric tensor on the manifold \cite{Palais_1957, wang2021GANGeometry}, and as such, it induces a norm on the latent space as $\norm{\mathbf{w}}^2_\mathbf{H} =  \mathbf{w}^\T\mathbf{H} \mathbf{w}$. 
It was shown in \cite{wang2021GANGeometry} that the Hessian $\mathbf{H}$ changes very little at different points on the latent space. 
Further, the eigenvalues of $\mathbf{H}$ have a fast decay and where only the top few eigenvectors induce a significant change in the generated images. 
For this reason, we can truncate the bottom eigenvectors leading to a speedup in training without affecting the editing quality.

\section{Experiments}

\subsection{Implementation Details}
We used the StyleGAN2 \cite{Karras2020StyleGAN2} networks pre-trained on FFHQ \cite{Karras2019StyleGAN} 
as well as StyleGAN3 pre-trained on FFHQU \cite{Karras2021StyleGAN3}. FFHQ is a data set of $7\times 10^4$ face images from flicker and FFHQU is the unaligned version. 
To construct the model $\mathcal{R}$ in \eqref{eq:rankonemodel} we first sampled $N=5\times10^4$ synthetic images and from each extracted $L = 68$ landmark points with Dlib \cite{dlib2009} and $L=468$ using MediaPipe \cite{Lugaresi2019MediaPipe}, which were then normalized to the interval $[0,1]$.
In each of the experiments, we have set the number of non-rigid basis shapes to $K=12$.
Further, we rotated the basis shapes to face the camera when $\bm{\theta} = \mathbf{0}$ in \eqref{eq:rankonemodel} in order to stabilize the training of the regressor.
We trained the regressor $\phi$, to predict the mean-centered output features $\hat{\mathbf{q}}$ for each of the $N$ samples. We used the Adam optimizer and a 10\% train vs. validation split, 5 hidden layers, each of size 512, and ReLU activation. 
To evaluate image similarity we use the Learned Perceptual Image Patch Similarity (LPIPS) \cite{Zhang2018LPIPS} and as a metric for identity similarity, we use Arcface \cite{Deng2019ArcFace}. 

\subsection{Model Evaluation}
\label{subsec:method_evaluation}
To evaluate our approach we randomly sampled $N = 1000$ samples $\mathbf{w}$ from the generator $G$ and measure the landmark loss
\begin{align}
    \mathcal{L}_\text{L}(\mathbf{w}) = ||(\mathcal{R} \circ \phi) ( \mathbf{w}) - (\psi_L \circ G)(\mathbf{w})||^2.
\end{align}
We then perform a series of edits $\mathbf{w}_\text{edit} = \Omega(\mathbf{w}, \mathbf{q}_\text{edit})$ on each sample of the linear baseline in \eqref{eq:linedit} 
For each edit, we  measure the landmark loss 
$ \mathcal{L}_\text{L}(\mathbf{w}_\text{edit})$ 
as well as three additional losses.
We measure how well the landmark prediction of the edited latent code agrees with the target ${R}(\mathbf{q}_\text{edit})$, i.e., how successfully we change the target attributes
\begin{align}
    \mathcal{L}_\phi = || \phi (\mathbf{w}_\text{edit}) - \mathbf{q}_\text{edit})||^2
 \end{align}
we measure how well the new GT landmarks align with the target
 \begin{align}
    \mathcal{L}_\mathcal{R} = || \mathcal{R}(\mathbf{q}_\text{edit}) - (\psi_L \circ G)(\mathbf{w}_\text{edit})||^2
 \end{align}
as well as the identity loss $ \mathcal{L}_\text{ID} (G(\mathbf{w}),G(\mathbf{w}_0))$, here \cite{Deng2019ArcFace}, between the original and edited image.

For this experiment, we used Mediapipe \cite{Lugaresi2019MediaPipe} as the landmark extractor $\psi_L$ and evaluated the full $1024^2$ resolution StyleGAN2 and 3 generators, both trained on the aligned FFHQ data set. The results can be seen in Table~\ref{table:evaluation}. 
The model $\phi$ had better performance in the $\mathcal{W}$ and $\mathcal{W}+$ spaces as compared to $\mathcal{Z}$ space when measuring the landmark loss $\mathcal{L}_L(\mathbf{w})$. This might be explained due to the fact the $\mathcal{Z}$ space is more entangled than $\mathcal{W}$ \cite{Karras2019StyleGAN}. 
We also observe that the identity loss $\mathcal{L}_{\text{ID}}$ is very low for $\mathcal{W}+$ space, however, $\mathcal{L}_\mathcal{R}$ is also dramatically higher, indicating that it is much harder to change the generated image in such a way that the extracted GT landmarks agree with the specified target when performing edits in $\mathcal{W}+$ space. The same point is supported by the $\mathcal{L}_\phi$ metric with is also substantially higher for $\mathcal{W}+$ space.  

\begin{table}[tb]
\caption{Model evaluation. 
Performance is measured across StyleGAN2 and 3 in $\mathcal{Z}$, $\mathcal{W}$, and $\mathcal{W}+$ space, respectively.}
\label{table:evaluation}
\begin{tabular}{c c c c c c } 
\toprule
$G$ &  $ \mathcal{L}_L(\mathbf{w})$  & $\mathcal{L}_L(\mathbf{w}_\text{edit})$  & $\mathcal{L}_\phi$ &  $\mathcal{L}_\mathcal{R}$ & $\mathcal{L}_\text{ID}$\\ 
\midrule
sg2 $\mathcal{Z}$ & 0.46 & 0.78 & 0.11 & 0.81 & 0.39\\ 
\hline
sg2 $\mathcal{W}$ &0.23 & 0.73 & 0.15 & 0.81 & 0.32\\ 
\hline
sg2 $\mathcal{W}+$ & 0.25 & 0.70 & 0.30 & 1.40 & 0.08\\ 
\hline
sg3 $\mathcal{Z}$ & 0.39 & 0.76 & 0.14 & 0.80 & 0.39\\ 
\hline
sg3 $\mathcal{W}$ & 0.26 & 0.56 & 0.19 & 0.66 & 0.36\\ 
\hline
sg3 $\mathcal{W}+$ & 0.25 & 0.71 & 0.28 & 1.28 & 0.18\\  
\bottomrule
\end{tabular}
\end{table}

\subsection{Identity Regularization}
\begin{figure}[tb]
\centering
\begin{flushleft}
\scriptsize 
\hspace{15pt} Original 
\hspace{35pt} Linear
\hspace{35pt} Non-linear 
\hspace{22pt} ${ \normalsize \substack{ \text{Non-linear}\\ \text{ w. ID loss} }}$ 
\end{flushleft}
\vspace{-10pt}
\includegraphics[ trim={0 0 0 0.4cm},clip, width=\linewidth]{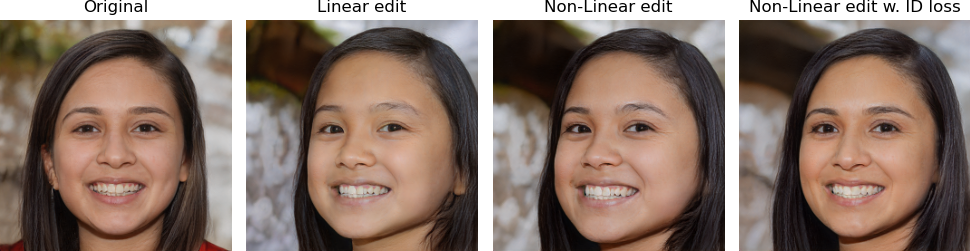}
\includegraphics[width=\linewidth]{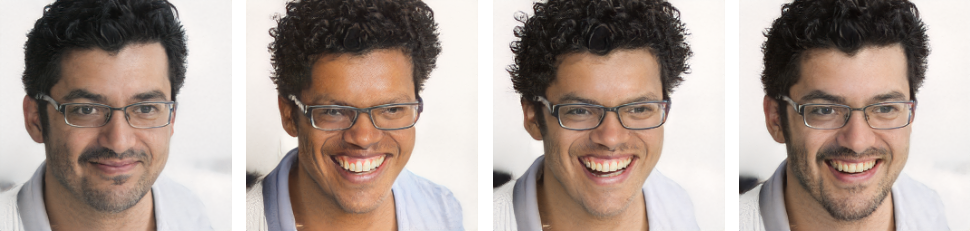}
\caption{The effect of adding identity regularization. We observe that adding ArcFace to the loss function improves the identity preservation of two edits: rotation (top) and smile (bottom).}
\label{fig:Regularization}
\end{figure}

We performed a qualitative comparison between the linear \eqref{eq:linedit} and gradient-based method \eqref{eq:gradientedit}, proposed in Section~\ref{subsec:editing}.
We visually compared editing pose and smile, using both methods, and visualized the effect of adding identity regularization using ArcFace \cite{Deng2019ArcFace} to the gradient-based method. The results can be seen in Figure~\ref{fig:Regularization}. 
We observe that the linear method is able to define directions in latent space which changes mostly the target attribute, here pose or smile, however, we note that the identity is not preserved well in the edit. 
This can be alleviated by the gradient-based method which defines a non-linear trajectory in latent space. 
Further, the gradient-based method in \eqref{eq:gradientedit} allows for explicit identity regularization using ArcFace \cite{Deng2019ArcFace} which substantially improves the degree of identity preservation for both pose and smile edits as can be seen in Figure~\ref{fig:Regularization}.

\subsection{Attribute Transfer}
\begin{figure}
\vspace{35pt}
\begin{picture}(200,200)
\put(0,0){\includegraphics[width=\linewidth]{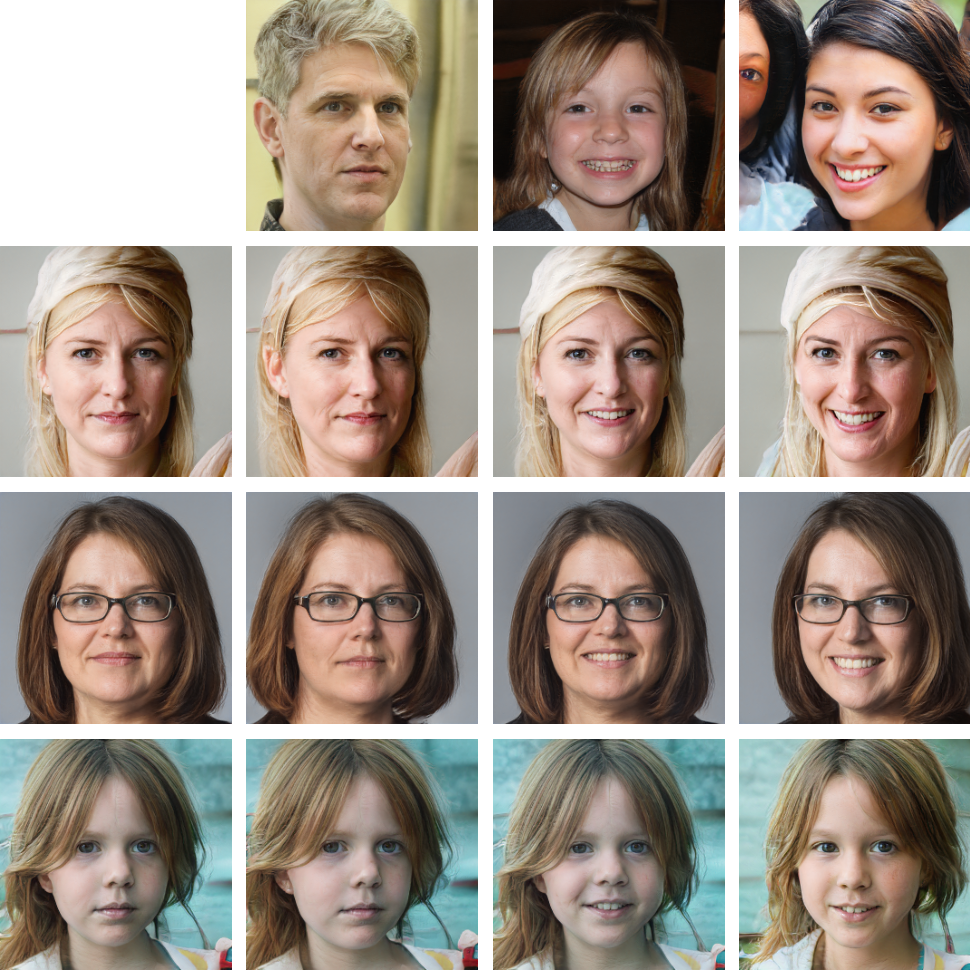}}
\put(15,180){Source}
\put(48,197){\rotatebox{90}{Target}}
\end{picture}
\caption{Our method allows for attribute transfer. We transfer rotation and expression from a source image and transfer these to a target image while preserving identity.}
\label{fig:transfergrid}
\end{figure}

Our approach enables attribute transfer in a straightforward manner. By attribute transfer, we refer to the process of transferring the pose and face shape from one image to another while preserving other attributes such as identity and illumination as much as possible. 
Given two latent codes, $\mathbf{w}_1$ and $\mathbf{w}_2$ with corresponding attribute vectors $\mathbf{q}_1$ and $\mathbf{q}_2$ we can transfer the pose and face shape from  $\mathbf{w}_1$ to  $\mathbf{w}_2$ by performing the edit $\tilde{\mathbf{w}}_2 = \Omega_{\mathbf{w}_2}(\mathbf{q}_1)$. Here both $\mathbf{q}_1$ and $\mathbf{q}_2$ can be recovered using either the regressor $\phi$ or using the minimization procedure in \eqref{eq:qrecover}.
In Figure~\ref{fig:transfergrid} we demonstrate attribute transfer by transferring the pose and face shape from 3 target images onto 3 source images. We see that the pose and face shapes are transferred while preserving the identity of the source images.

%

\subsection{Rotation and Translation with StyleGAN3}
\begin{figure}
{\scriptsize \hspace{20pt} Original \hspace{26pt} Translation \hspace{25pt} Rotation \hspace{12pt} {Rotation+Translation}} \\
\rotatebox{90}{\hspace{10pt}{\scriptsize SG3-native}} 
\includegraphics[width=0.45\textwidth]{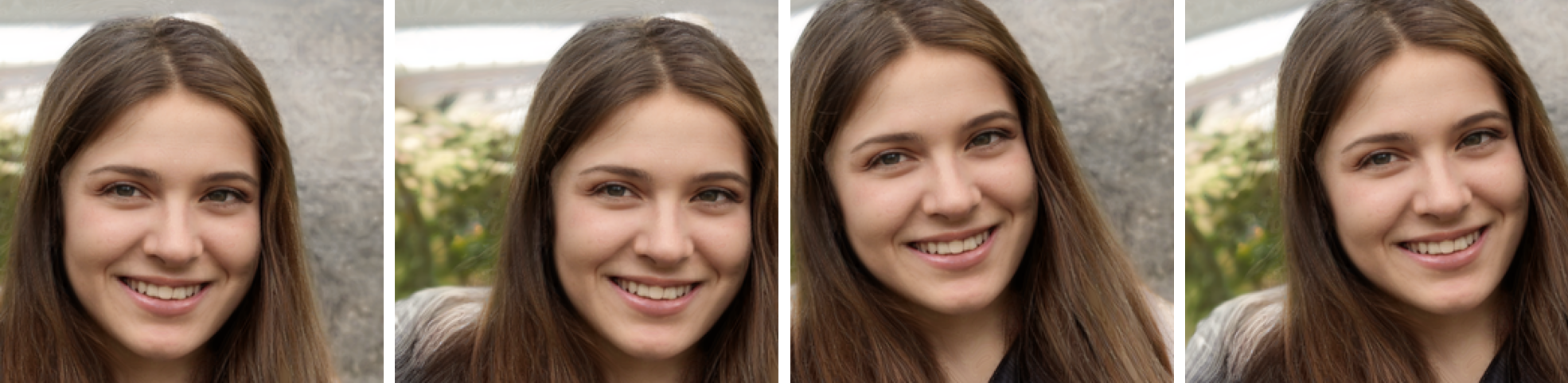}\\
\rotatebox{90}{\hspace{20pt}{\scriptsize Ours}} 
\includegraphics[width=0.45\textwidth]{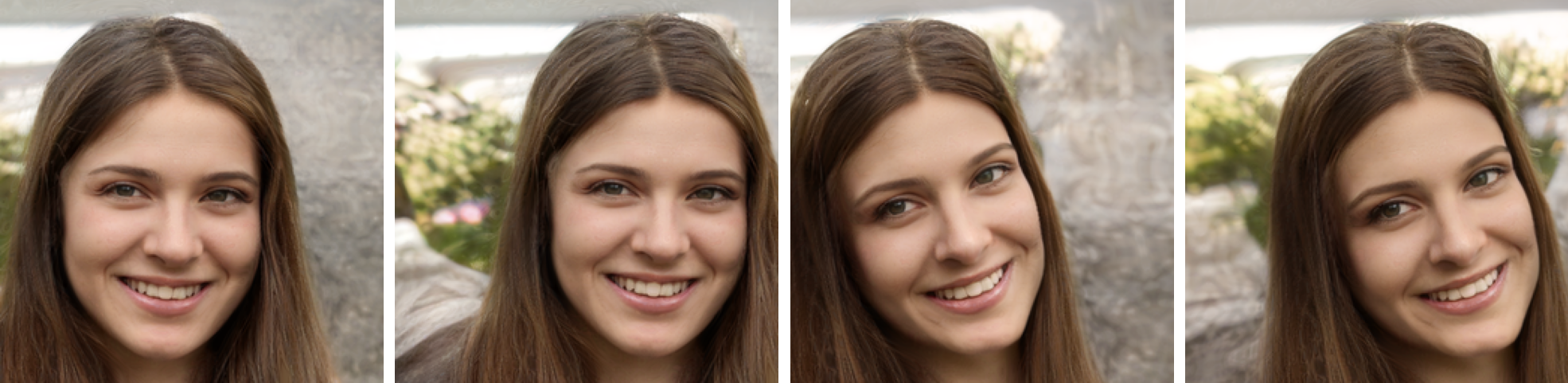}
\caption{The effect of performing translation, roll rotation or both with our method compared to directly editing the Fourier features in StyleGAN3.}
\label{fig:exp2-sg3-roll-trans}
\end{figure}

Our method is able to define trajectories in latent space corresponding to roll rotation as well as translations. 
As noted in \cite{Alaluf2022SGThirdTime} roll rotations and translations are a native part of the architecture of the StyleGAN3 \cite{Karras2021StyleGAN3} generator by manipulating Fourier features with the four parameters $ (\sin \alpha,\cos \alpha , x, y)$ which are obtained from the first learned affine layer of the synthesis network.
We qualitatively compared the effect of performing roll rotation and translation using our method to the effect of manipulating the Fourier features directly. The results can be seen in Figure~\ref{fig:exp2-sg3-roll-trans}.
We note translations look very similar with both methods.
However, for roll rotations, we note that the axis of rotation is located in the middle of the left-hand image border when manipulating the Fourier features (see the location of the nose in Figure~\ref{fig:exp2-sg3-roll-trans}), whereas, with our method, the axis of rotation is located at the center of the face.


\subsection{Comparison with other methods}
\begin{figure}
\centering
\begin{subfigure}[b]{\linewidth}
\centering
\begin{flushleft}
\scriptsize 
\hspace{10pt} Original 
\hspace{18pt} GANSpace
\hspace{6pt} InterFaceGAN 
\hspace{6pt} TensorGAN 
\hspace{19pt} Ours
\end{flushleft}
\vspace{-10pt}
\includegraphics[width=\linewidth]{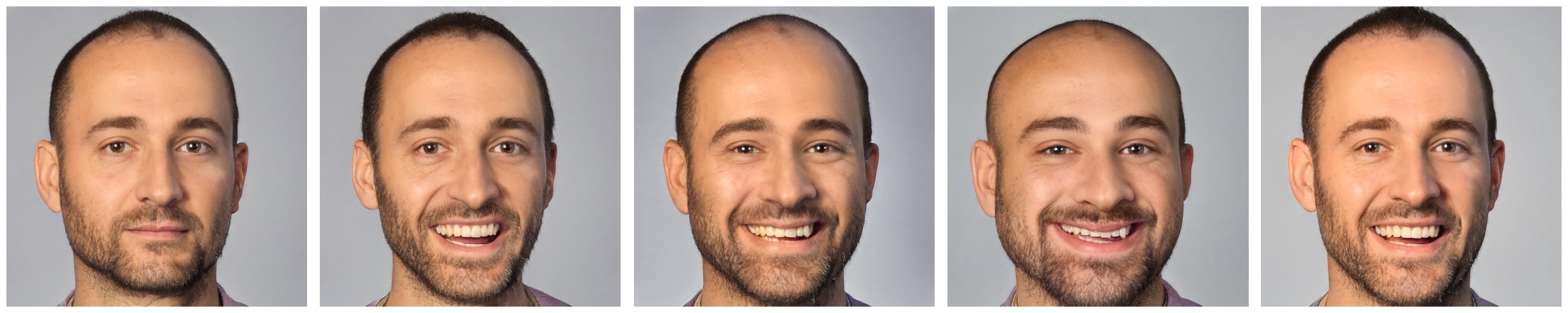}
\includegraphics[width=\linewidth]{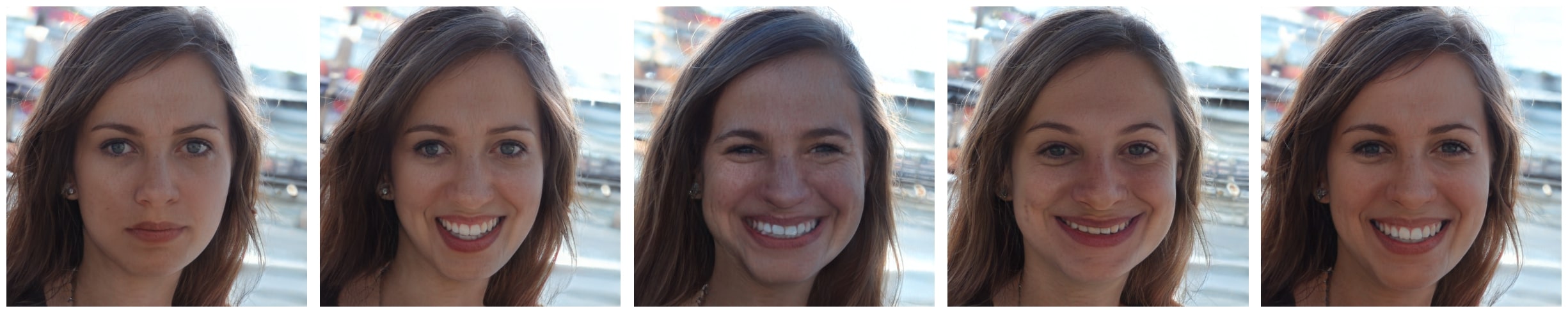}
\includegraphics[width=\linewidth]{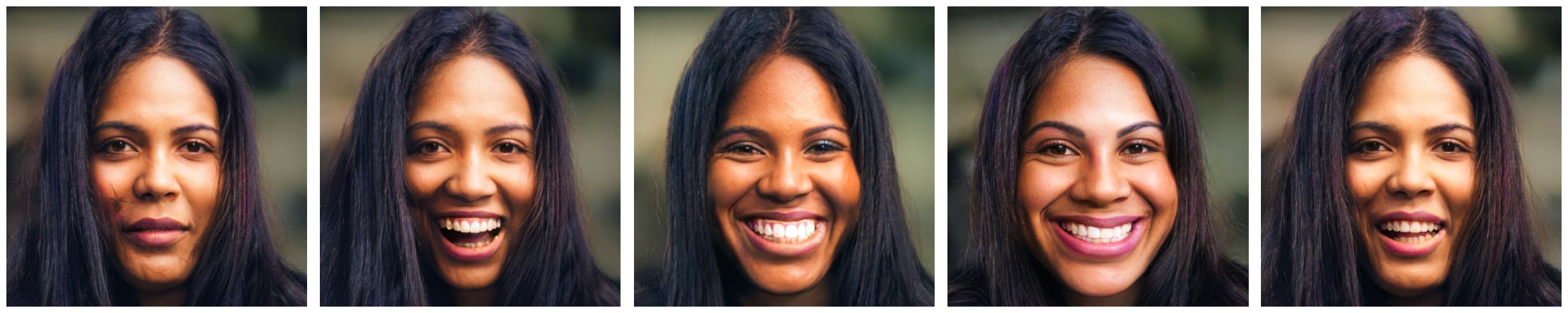}
\caption{Smile edits}
\label{fig:exp1-smile-compare}
\end{subfigure}

\begin{subfigure}[b]{\linewidth}
\rotatebox{90}{
\hspace{5pt}{\scriptsize GANSpace}
} 
\hspace{-1.3pt}
\includegraphics[width=\linewidth]{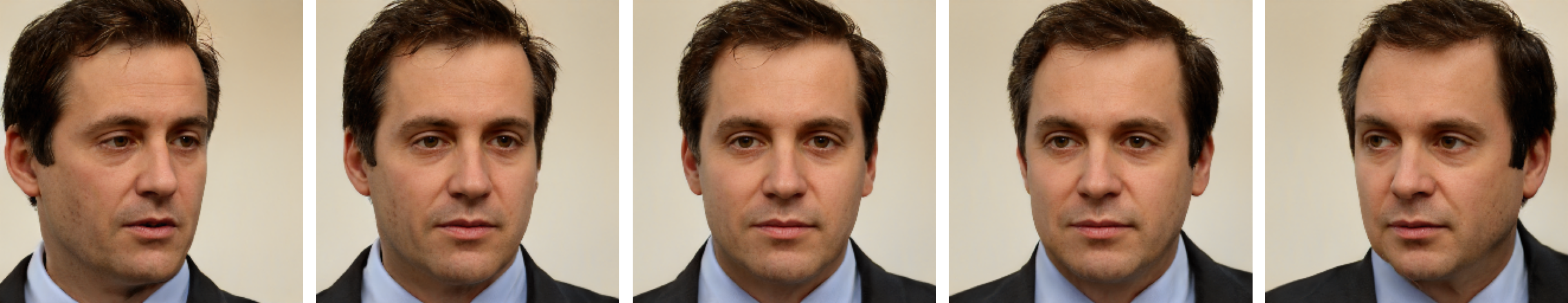}
\rotatebox{90}{\hspace{.0pt} {\scriptsize InterFaceGAN}}
\hspace{0.1pt}
\includegraphics[width=\linewidth]{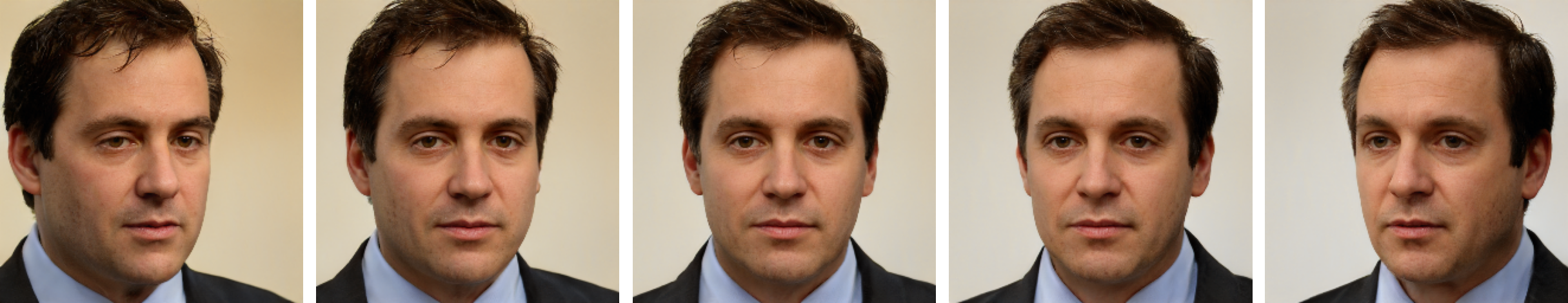}
\rotatebox{90}{\hspace{2pt} {\scriptsize TensorGAN}}
\hspace{0.1pt}
\includegraphics[width=\linewidth]{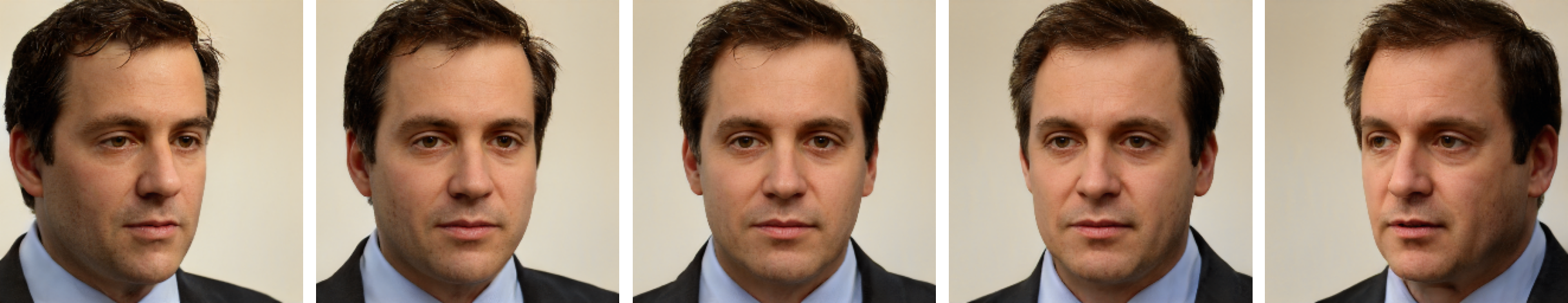}
\rotatebox{90}{\hspace{12pt} {\scriptsize Ours}} 
\hspace{0.1pt}
\includegraphics[width=\linewidth]{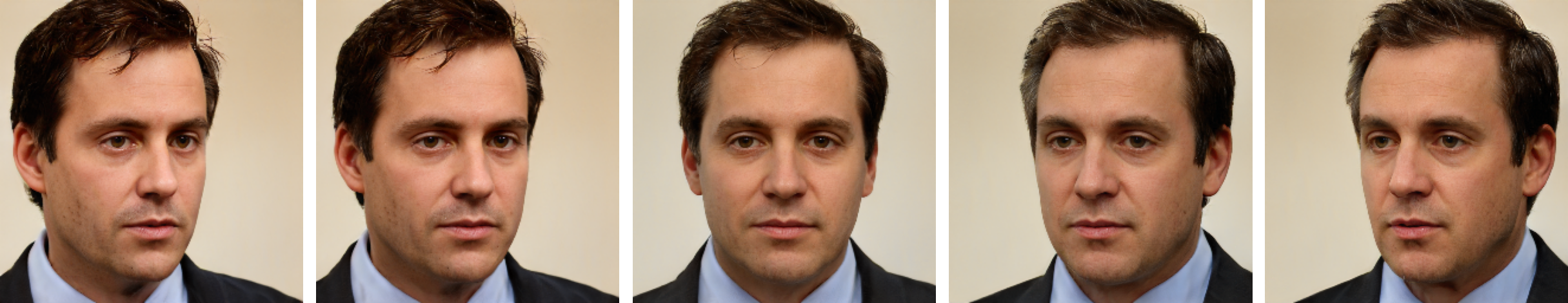}
\caption{Yaw rotation edits}
\label{fig:exp1-yaw-rot-compare}
\end{subfigure}
\caption{Qualitative comparison of editing smile and yaw rotation and using our method with other off-the-shelf techniques.}
\end{figure}

We compared the editing directions corresponding to pose (yaw rotation) and smile with three off-the-shelf techniques for semantic editing: InterFaceGAN \cite{Shen2020Interfacegan,Shen2020InterfaceganTPAMI}, GANSpace \cite{Harkonen2020GANSpace}, and TensorGAN \cite{Haas2021tensorGAN, Haas2022tensorGAN2}.
Although our method supports arbitrary 3D rotations in latent space, we focused on editing yaw rotations and smile since previous techniques have also been reported to enable these edits, enabling a direct comparison. 
A qualitative comparison of the edits to smile and yaw rotations generated by each of the four methods is shown in Figure~\ref{fig:exp1-yaw-rot-compare}. %
When evaluating the degree of identity preservation during the semantic edits it can be seen that our method is on par with the competing methods when performing yaw rotations and arguably better when editing smile.

\subsection{Editing real images}
\begin{figure}
    \begin{flushleft}
    \small 
    \hspace{14pt} Real Image 
    \hspace{7pt} Reconstruction
    \hspace{11pt} Pose Edit 
    \hspace{19pt} Smile edit 
    \end{flushleft}
    \vspace{-10pt}
    \rotatebox{90}{\hspace{25pt}{\scriptsize e4e}} 
    \includegraphics[width=\linewidth]{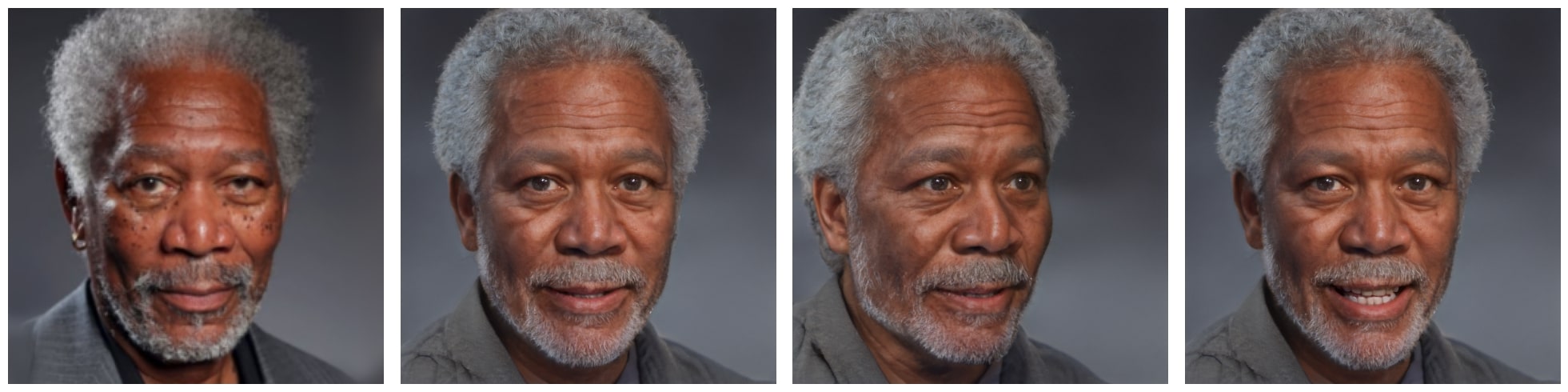}
    
    \rotatebox{90}{\hspace{13pt}{\scriptsize HyperStyle}} 
    \hspace{-4pt}
    \includegraphics[width=\linewidth]{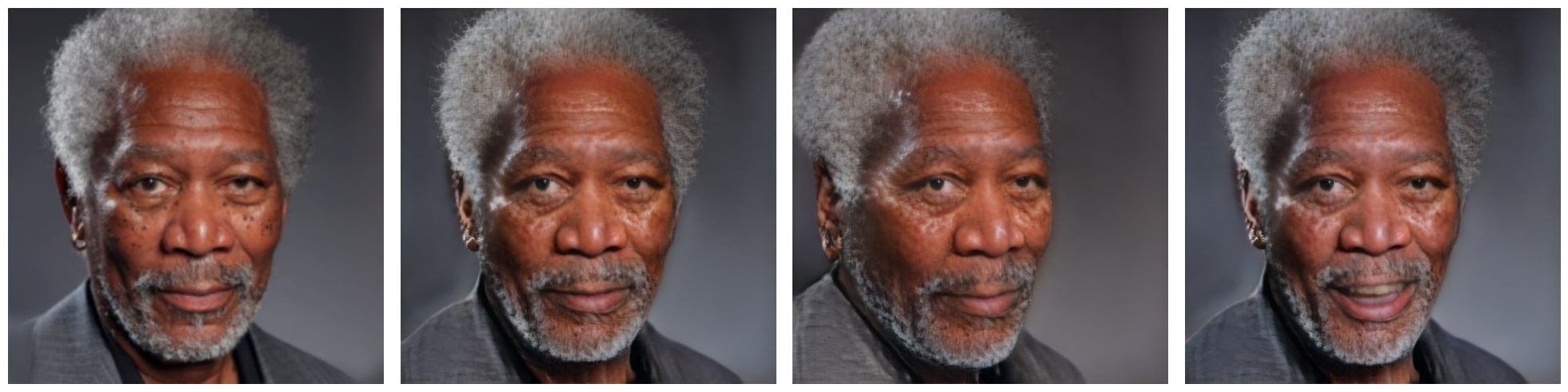}
    \caption{Qualitative comparison of projection and editing results when combining our method with two state-of-the-art encoders, e4e \cite{Tov2021e4e} and HyperStyle \cite{alaluf2021hyperstyle} respectively. }
    \label{fig:real_images}
    \end{figure}

Coupled with an encoder, our approach facilitates the editing of real images. 
We qualitatively compared the projection and editing results when using our method in conjunction with e4e \cite{Tov2021e4e} and HyperStyle \cite{alaluf2021hyperstyle}, respectively. 
The results can be seen in Figure~\ref{fig:real_images}. 
The two methods operate in different spaces, e4e project images into $\mathcal{W}+$ space while HyperStyle instead makes an initial prediction in $\mathcal{W}$ space and then fine-tunes the generator such that the prediction more faithfully reconstructs the target. Despite the fine-tuning of the generator it is not necessary to retrain the regressor $\phi$ when using HyperStyle for GAN Inversion. 

\section{Conclusions}
We have presented a framework for highly interpretable image editing in pre-trained 2D GANs. Our framework provides an efficient method to find trajectories in the latent space of pre-trained generators which change the generated images according to camera, orientation, and shape parameters. 
With this, we can define trajectories in latent space corresponding to arbitrary transformations of shape and orientation of the generated images. 
Moreover, since our approach allows for the modeling of dense shapes, this work can be seen as an extension of NRSfM for generative models.

To summarize our approach, we first used NRSfM to derive a sparse 3D model on the domain of the generator. We then trained a regressor to relate the 3D model to the latent space of a 2D GAN. We proposed two methods for using the regressor for semantic editing, a linear method, and a gradient-based method. Our gradient-based method is similar to the iterative editing algorithm proposed in \cite{wang2021hijackgan}, however, in our work we integrate explicit identity regularization which improves identity preservation when editing face images.  

Our method provides an efficient framework for manipulating the 3D structure of objects generated by 2D GANs. Compared to other methods, our approach is fast compared to existing frameworks for training explicitly 3D aware GANs \cite{gu2021stylenerf,Niemeyer2021giraffe,ylmaz2022liftedgan} and compared to \cite{Tewari2020StyleRig} our method is lightweight and able to perform rotations and edits to face shape without the need for a 3D morphable model.
Since our method only requires access to a landmark extractor trained on the same domain as the generator, our approach does not require any additional training data and can be trained in a fully self-supervised fashion. Further, our approach does not require retraining of the generator or any changes to the generator architecture. 

As to limitations, our method allows for adjustments to the position, orientation as well as non-rigid deformation of the face shape of the generated images. Since our method only captures the 3D orientation and face shape our method is not able to add or remove face accessories, eye-glasses, earrings, and hats nor change the skin tone or hair color. 
Exploring how to extend our NRSfM model to overcome these limitations is an interesting avenue for future work. 

\section{Ethics Statement}
This work proposes a novel framework for combining NRSfM with generative models. Coupled with a latent encoder, our method is able to edit the pose and non-rigid face shape of real face images. As our method relies on a pre-trained GAN, it inherits the biases which might be present in the training data of the input GAN. 
Generative models, particularly models which are trained on human faces, raise several concerns as they can be used to generate manipulative or offensive content such as deepfakes. However, these concerns apply to the field at large and are not exacerbated by this work. 




\newpage
{\small
\bibliographystyle{ieee_fullname}
\bibliography{references.bib}

\begin{thebibliography}{10}\itemsep=-1pt

\bibitem{abdal2020img2sg++}
Rameen Abdal, Yipeng Qin, and Peter Wonka.
\newblock Image2stylegan++: How to edit the embedded images?
\newblock In {\em 2020 IEEE/CVF Conference on Computer Vision and Pattern
  Recognition (CVPR)}, page 8293–8302, Seattle, WA, USA, Jun 2020. IEEE.

\bibitem{Abdal2020StyleFlow}
Rameen Abdal, Peihao Zhu, Niloy~J. Mitra, and Peter Wonka.
\newblock Styleflow: Attribute-conditioned exploration of stylegan-generated
  images using conditional continuous normalizing flows.
\newblock {\em ACM Trans. Graph.}, 40(3), May 2021.

\bibitem{alaluf2021restyle}
Yuval Alaluf, Or Patashnik, and Daniel Cohen-Or.
\newblock Restyle: A residual-based stylegan encoder via iterative refinement.
\newblock In {\em Proceedings of the IEEE/CVF International Conference on
  Computer Vision (ICCV)}, October 2021.

\bibitem{Alaluf2022SGThirdTime}
Yuval Alaluf, Or Patashnik, Zongze Wu, Asif Zamir, Eli Shechtman, Dani
  Lischinski, and Daniel Cohen-Or.
\newblock Third time’s the charm? image and video editing with stylegan3.
\newblock {\em Advances in Image Manipulation Workshop - ECCV 2022}, Jan 2022.

\bibitem{alaluf2021hyperstyle}
Yuval Alaluf, Omer Tov, Ron Mokady, Rinon Gal, and Amit~H. Bermano.
\newblock Hyperstyle: Stylegan inversion with hypernetworks for real image
  editing.
\newblock In {\em Proc.\@ CVPR}, 2022.

\bibitem{pbaylies}
Peter Baylies.
\newblock Stylegan encoder - converts real images to latent space.
\newblock \url{https://github.com/pbaylies/styleganencoder/}, 2019.

\bibitem{Blanz1999MorphableModel}
Volker Blanz and Thomas Vetter.
\newblock A morphable model for the synthesis of {3D} faces.
\newblock In {\em Proc.\@ SIGGRAPH}, pages 187--194, 1999.

\bibitem{Brandt_Ackermann_2019}
Sami~S. Brandt and Hanno Ackermann.
\newblock Non-rigid structure-from-motion by rank-one basis shapes, Apr 2019.
\newblock arXiv: 1904.13271.

\bibitem{Brandt_Ackermann_Grasshof}
Sami~Sebastian Brandt, Hanno Ackermann, and Stella Grasshof.
\newblock Uncalibrated non-rigid factorisation by independent subspace
  analysis.
\newblock In {\em 2019 IEEE/CVF International Conference on Computer Vision
  Workshop (ICCVW)}, pages 569--578, 2019.

\bibitem{Bregler_Hertzmann_Biermann_2000}
C. Bregler, A. Hertzmann, and H. Biermann.
\newblock Recovering non-rigid 3d shape from image streams.
\newblock In {\em Proceedings IEEE Conference on Computer Vision and Pattern
  Recognition. CVPR 2000 (Cat. No.PR00662)}, volume~2, page 690–696. IEEE
  Comput. Soc, 2000.

\bibitem{choi2018stargan}
Yunjey Choi, Minje Choi, Munyoung Kim, Jung-Woo Ha, Sunghun Kim, and Jaegul
  Choo.
\newblock Stargan: Unified generative adversarial networks for multi-domain
  image-to-image translation.
\newblock In {\em Proceedings of the IEEE Conference on Computer Vision and
  Pattern Recognition}, 2018.

\bibitem{choi2020starganv2}
Yunjey Choi, Youngjung Uh, Jaejun Yoo, and Jung-Woo Ha.
\newblock Stargan v2: Diverse image synthesis for multiple domains.
\newblock In {\em Proceedings of the IEEE Conference on Computer Vision and
  Pattern Recognition}, 2020.

\bibitem{Deng2019ArcFace}
Jiankang Deng, Jia Guo, Niannan Xue, and Stefanos Zafeiriou.
\newblock {Arcface}: {A}dditive angular margin loss for deep face recognition.
\newblock In {\em Proc.\@ CVPR}, pages 4690--4699, 2019.

\bibitem{Gatys2015StyleTransfer}
Leon~A. Gatys, Alexander~S. Ecker, and Matthias Bethge.
\newblock A neural algorithm of artistic style.
\newblock {\em CoRR}, abs/1508.06576, 2015.

\bibitem{Goodfellow2014GAN}
Ian Goodfellow, Jean Pouget-Abadie, Mehdi Mirza, Bing Xu, David Warde-Farley,
  Sherjil Ozair, Aaron Courville, and Yoshua Bengio.
\newblock Generative adversarial nets.
\newblock In Z. Ghahramani, M. Welling, C. Cortes, N. Lawrence, and K.~Q.
  Weinberger, editors, {\em Advances in Neural Information Processing Systems},
  volume~27, page 2672–2680. Curran Associates, Inc., 2014.

\bibitem{Grasshof_Brandt_2022}
Stella Graßhof and Sami~Sebastian Brandt.
\newblock Tensor-based non-rigid structure from motion.
\newblock In {\em 2022 IEEE/CVF Winter Conference on Applications of Computer
  Vision (WACV)}, page 2254–2263. IEEE, Jan 2022.

\bibitem{gu2021stylenerf}
Jiatao Gu, Lingjie Liu, Peng Wang, and Christian Theobalt.
\newblock Stylenerf: A style-based 3d aware generator for high-resolution image
  synthesis.
\newblock In {\em International Conference on Learning Representations}, 2022.

\bibitem{Haas2021tensorGAN}
René Haas, Stella Graßhof, and Sami~Sebastian Brandt.
\newblock Tensor-based subspace factorization for stylegan.
\newblock In {\em 2021 16th IEEE International Conference on Automatic Face and
  Gesture Recognition (FG 2021)}, pages 1--8, Los Alamitos, CA, USA, dec 2021.
  IEEE Computer Society.

\bibitem{Haas2022tensorGAN2}
René Haas, Stella Graßhof, and Sami~Sebastian Brandt.
\newblock Tensor-based emotion editing in the stylegan latent space.
\newblock {\em arXiv:2205.06102 [cs]}, May 2022.
\newblock Accepted for poster presentation at AI4CC @ CVPRW.

\bibitem{Huang2017StyleTransfer}
Xun Huang and Serge Belongie.
\newblock Arbitrary style transfer in real-time with adaptive instance
  normalization.
\newblock {\em Proc.\@ ICCV}, Jul 2017.

\bibitem{Harkonen2020GANSpace}
Erik Härkönen, Aaron Hertzmann, Jaakko Lehtinen, and Sylvain Paris.
\newblock Ganspace: Discovering interpretable gan controls.
\newblock In {\em Proc.\@ NeurIPS}, 2020.

\bibitem{bue2021nrsfmbenchmark}
Sebastian Hoppe~Nesgaard Jensen, Mads Emil~Brix Doest, Henrik Aanæs, and
  Alessio Del~Bue.
\newblock A benchmark and evaluation of non-rigid structure from motion.
\newblock {\em International Journal of Computer Vision}, 129(4):882–899, Apr
  2021.

\bibitem{karras2018pggan}
Tero Karras, Timo Aila, Samuli Laine, and Jaakko Lehtinen.
\newblock Progressive growing of gans for improved quality, stability, and
  variation.
\newblock In {\em Proc.\@ ICLR}, Feb 2018.

\bibitem{Karras2020StyleGANada}
Tero Karras, Miika Aittala, Janne Hellsten, Samuli Laine, Jaakko Lehtinen, and
  Timo Aila.
\newblock Training generative adversarial networks with limited data.
\newblock In {\em Proc.\@ NeurIPS}, 2020.

\bibitem{Karras2021StyleGAN3}
Tero Karras, Miika Aittala, Samuli Laine, Erik H\"ark\"onen, Janne Hellsten,
  Jaakko Lehtinen, and Timo Aila.
\newblock Alias-free generative adversarial networks.
\newblock In {\em Proc.\@ NeurIPS}, 2021.

\bibitem{Karras2019StyleGAN}
Tero Karras, Samuli Laine, and Timo Aila.
\newblock A style-based generator architecture for generative adversarial
  networks.
\newblock In {\em Proc.\@ CVPR}, pages 4396--4405, 2019.

\bibitem{Karras2020StyleGAN2}
Tero Karras, Samuli Laine, Miika Aittala, Janne Hellsten, Jaakko Lehtinen, and
  Timo Aila.
\newblock Analyzing and improving the image quality of {StyleGAN}.
\newblock In {\em Proc.\@ CVPR}, 2020.

\bibitem{dlib2009}
Davis~E. King.
\newblock Dlib-ml: {A} {Machine} {Learning} {Toolkit}.
\newblock {\em J. Mach. Learn. Res.}, 10:1755--1758, Dec. 2009.

\bibitem{KongLucey2019DeepNRSFM}
Chen Kong and Simon Lucey.
\newblock Deep {Non}-{Rigid} {Structure} {From} {Motion}.
\newblock In {\em 2019 {IEEE}/{CVF} {International} {Conference} on {Computer}
  {Vision} ({ICCV})}, pages 1558--1567, Oct. 2019.
\newblock ISSN: 2380-7504.

\bibitem{Lugaresi2019MediaPipe}
Camillo Lugaresi, Jiuqiang Tang, Hadon Nash, Chris McClanahan, Esha Uboweja,
  Michael Hays, Fan Zhang, Chuo-Ling Chang, Ming~Guang Yong, Juhyun Lee, and et
  al.
\newblock Mediapipe: A framework for building perception pipelines.
\newblock {\em arXiv:1906.08172 [cs]}, Jun 2019.
\newblock arXiv: 1906.08172.

\bibitem{Niemeyer2021giraffe}
Michael Niemeyer and Andreas Geiger.
\newblock Giraffe: Representing scenes as compositional generative neural
  feature fields.
\newblock In {\em 2021 IEEE/CVF Conference on Computer Vision and Pattern
  Recognition (CVPR)}, page 11448–11459, Nashville, TN, USA, Jun 2021. IEEE.

\bibitem{nitzan2021large}
Yotam Nitzan, Rinon Gal, Ofir Brenner, and Daniel Cohen-Or.
\newblock Large: Latent-based regression through gan semantics.
\newblock In {\em 2022 IEEE/CVF Conference on Computer Vision and Pattern
  Recognition (CVPR)}, pages 19217--19227, 2022.

\bibitem{Palais_1957}
Richard Palais.
\newblock On the differentiability of isometries.
\newblock {\em Proceedings of The American Mathematical Society - PROC AMER
  MATH SOC}, 8:805–805, Aug 1957.

\bibitem{patashnik2021styleclip}
Or Patashnik, Zongze Wu, Eli Shechtman, Daniel Cohen-Or, and Dani Lischinski.
\newblock Styleclip: Text-driven manipulation of stylegan imagery.
\newblock In {\em Proceedings of the IEEE/CVF International Conference on
  Computer Vision (ICCV)}, pages 2085--2094, October 2021.

\bibitem{Perarnau2016Invertible}
Guim Perarnau, Joost van~de Weijer, Bogdan Raducanu, and Jose~M. Álvarez.
\newblock Invertible conditional gans for image editing.
\newblock In {\em NIPS 2016 Workshop on Adversarial Training}, Nov 2016.

\bibitem{Abdal2019Image2StyleGAN}
Yipeng~Qin Rameen~Abdal and Peter Wonka.
\newblock Image2stylegan: How to embed images into the stylegan latent space?
\newblock In {\em Proc.\@ ICCV}, pages 4431--4440, 2019.

\bibitem{richardson2021encoding}
Elad Richardson, Yuval Alaluf, Or Patashnik, Yotam Nitzan, Yaniv Azar, Stav
  Shapiro, and Daniel Cohen-Or.
\newblock Encoding in style: a stylegan encoder for image-to-image translation.
\newblock In {\em IEEE/CVF Conference on Computer Vision and Pattern
  Recognition (CVPR)}, June 2021.

\bibitem{Roich2021pivotal}
Daniel Roich, Ron Mokady, Amit~H Bermano, and Daniel Cohen-Or.
\newblock Pivotal tuning for latent-based editing of real images.
\newblock {\em ACM Trans. Graph.}, 2021.

\bibitem{Shao2018RiemannDGM}
Hang Shao, Abhishek Kumar, and P.~Thomas Fletcher.
\newblock The riemannian geometry of deep generative models.
\newblock In {\em 2018 IEEE/CVF Conference on Computer Vision and Pattern
  Recognition Workshops (CVPRW)}, page 428–4288, Salt Lake City, UT, USA, Jun
  2018. IEEE.

\bibitem{Shen2020Interfacegan}
Yujun Shen, Jinjin Gu, Xiaoou Tang, and Bolei Zhou.
\newblock Interpreting the latent space of gans for semantic face editing.
\newblock In {\em CVPR}, 2020.

\bibitem{Shen2020InterfaceganTPAMI}
Yujun Shen, Ceyuan Yang, Xiaoou Tang, and Bolei Zhou.
\newblock Interfacegan: Interpreting the disentangled face representation
  learned by gans.
\newblock {\em TPAMI}, 2020.

\bibitem{Shen2020SeFa}
Yujun Shen and Bolei Zhou.
\newblock Closed-form factorization of latent semantics in gans.
\newblock In {\em CVPR}, 2021.

\bibitem{sidhu_neural_2020}
Vikramjit Sidhu, Edgar Tretschk, Vladislav Golyanik, Antonio Agudo, and
  Christian Theobalt.
\newblock Neural {Dense} {Non}-{Rigid} {Structure} from {Motion} with {Latent}
  {Space} {Constraints}.
\newblock In {\em European Conference on Computer Vision (ECCV)}, 2020.

\bibitem{Tewari2020StyleRig}
Ayush Tewari, Mohamed Elgharib, Gaurav Bharaj, Florian Bernard, Hans-Peter
  Seidel, Patrick P{\'e}rez, Michael Z{\"o}llhofer, and Christian Theobalt.
\newblock Stylerig: Rigging stylegan for 3d control over portrait images, cvpr
  2020.
\newblock In {\em Proc.\@ CVPR)}. {IEEE}, june 2020.

\bibitem{Tomasi_Kanade_1992}
Carlo Tomasi and Takeo Kanade.
\newblock Shape and motion from image streams under orthography: a
  factorization method.
\newblock {\em International Journal of Computer Vision}, 9(2):137–154, 1992.

\bibitem{Tov2021e4e}
Omer Tov, Yuval Alaluf, Yotam Nitzan, Or Patashnik, and Daniel Cohen-Or.
\newblock Designing an encoder for {StyleGAN} image manipulation.
\newblock {\em ACM Transactions on Graphics}, 40(4):133:1--133:14, July 2021.

\bibitem{wang2021GANGeometry}
Binxu Wang and Carlos~R Ponce.
\newblock A geometric analysis of deep generative image models and its
  applications.
\newblock In {\em International Conference on Learning Representations}, 2021.

\bibitem{wang2021hijackgan}
Hui-Po Wang, Ning Yu, and Mario Fritz.
\newblock Hijack-gan: Unintended-use of pretrained, black-box gans.
\newblock In {\em Proc.\@ CVPR)}, page~10, 2021.

\bibitem{Wu2020StyleSpace}
Zongze Wu, Dani Lischinski, and Eli Shechtman.
\newblock Stylespace analysis: Disentangled controls for stylegan image
  generation.
\newblock In {\em Proc.\@ CVPR}, Dec 2020.

\bibitem{ylmaz2022liftedgan}
Do{\u{g}}a Y{\i}lmaz, Furkan K{\i}nl{\i}, Bar{\i}{\c{s}} {\"O}zcan, and Furkan
  K{\i}ra{\c{c}}.
\newblock [re] lifting 2d style{GAN} for 3d-aware face generation.
\newblock In {\em ML Reproducibility Challenge 2021 (Fall Edition)}, 2022.

\bibitem{Zhang2018LPIPS}
Richard Zhang, Phillip Isola, Alexei~A Efros, Eli Shechtman, and Oliver Wang.
\newblock The unreasonable effectiveness of deep features as a perceptual
  metric.
\newblock In {\em Proc\@ CVPR}, 2018.

\bibitem{Zhu2020InDomain}
Jiapeng Zhu, Yujun Shen, Deli Zhao, and Bolei Zhou.
\newblock In-domain gan inversion for real image editing.
\newblock In {\em Proc. \@ ECCV}, 2020.

\end{thebibliography}
}


\end{document}